\documentclass[10pt,twocolumn,letterpaper]{article}

\usepackage[pagenumbers]{cvpr} %
\usepackage[dvipsnames]{xcolor}

\usepackage{amsmath}
\usepackage{amssymb}
\usepackage{algorithmic}
\usepackage{algorithm}
\usepackage{amsthm}
\usepackage{mathtools}

\usepackage{tabularx}
\usepackage{array}
\usepackage{makecell}
\usepackage{multicol}
\usepackage{multirow}
\usepackage{booktabs} %
\usepackage{tcolorbox}
\usepackage{pifont}
\newcommand{\summary}[1]{
\begin{center}
\begin{tcolorbox}[colback=gray!15, colframe=black, boxsep=-0.15cm, middle=-0.15cm]
\textbf{\ding{46} Guidance}
$\blacktriangleright$
{#1}
$\blacktriangleleft$
\end{tcolorbox}
\end{center}
}

\usepackage{amsmath,amsfonts,bm}

\def\eqref#1{equation~\ref{#1}}

\def\1{\bm{1}}

\def\vc{{\bm{c}}}

\def\ve{{\bm{e}}}

\def\vg{{\bm{g}}}

\def\vp{{\bm{p}}}

\def\vt{{\bm{t}}}

\def\vx{{\bm{x}}}

\DeclareMathAlphabet{\mathsfit}{\encodingdefault}{\sfdefault}{m}{sl}
\SetMathAlphabet{\mathsfit}{bold}{\encodingdefault}{\sfdefault}{bx}{n}

\def\gX{{\mathcal{X}}}

\newcommand{\ours}{\textsc{CCRT}}

\newcommand{\circled}[1]{\textcircled{\raisebox{-.9pt}{#1}}}

\definecolor{cvprblue}{rgb}{0.21,0.49,0.74}
\usepackage[pagebackref,breaklinks,colorlinks,allcolors=cvprblue]{hyperref}

\usepackage{makecell}

\def\paperID{11038} %
\def\confName{CVPR}
\def\confYear{2025}
\definecolor{cvprblue}{rgb}{0.21,0.49,0.74}
\title{Continuous Concepts Removal in Text-to-image Diffusion Models}

\author{$\text{Tingxu Han}^1$,$ \text{Weisong Sun}^2$, $\text{Yanrong Hu}^3$, $\text{Chunrong Fang}^1$, $\text{Yonglong Zhang}^3$, \\
$\text{Shiqing Ma}^4$, $\text{Tao Zheng}^1$, $\text{Zhenyu Chen}^1$, $\text{Zhenting Wang}^5$ \\
$^1\text{Nanjing University}$,
$^2\text{Nanyang Technological University}$, $^3\text{Yangzhou University}$, \\
$^4\text{University of Massachusetts at Amherst}$, $^5\text{Rutgers University}$\\
{\tt\small txhan@smail.nju.edu.cn, weisong.sun@ntu.edu.sg, 201404312@stu.yzu.edu.cn,}\\
{\tt\small \{fangchunrong, zt, zychen\}@nju.edu.cn, ylzhang@yzu.edu.cn} \\
{\tt\small shiqingma@umass.edu, zhenting.wang@rutgers.edu} \\
}

\begin{document}
\maketitle
\
\begin{abstract}

Text-to-image diffusion models have shown an impressive ability to generate high-quality images from input textual descriptions. However, concerns have been raised about the potential for these models to create content that infringes on copyrights or depicts disturbing subject matter.
Removing specific concepts from these models is a promising potential solution to this problem. However, existing methods for concept removal do not work well in practical but challenging scenarios where concepts need to be continuously removed. Specifically, these methods lead to poor alignment between the text prompts and the generated image after the continuous removal process.
To address this issue, we propose a novel approach called \ours{} that includes a designed knowledge distillation paradigm.
It constrains the text-image alignment behavior during the continuous concept removal process by using a set of text prompts generated through our genetic algorithm, which employs a designed fuzzing strategy. 
We conduct extensive experiments involving the removal of various concepts.
The results evaluated through both algorithmic metrics and human studies demonstrate that our \ours{} can effectively remove the targeted concepts in a continuous manner while maintaining the high generation quality (e.g., text-image alignment) of the model.

\end{abstract}

\vspace{-12pt}
\section{Introduction}
\label{sec:intro}

Advancements in Artificial Intelligence Generated Contents (AIGCs)~\cite{wu2023ai} have revolutionized the field of image synthesis~\cite{rombach2022high, song2023consistency, xue2024raphael}, among which text-to-image diffusion models enable the creation of high-quality images from textual descriptions~\cite{saharia2022photorealistic, zhang2023adding}. 
However, this progress has also raised significant concerns regarding the potential misuse of these models~\cite{ghosh2022can, carlini2023extracting, wang2023diagnosis, sha2023fake}. 
Such misuse includes generating content that infringes on copyrights, such as mimicking specific artistic styles~\cite{roose2022ai}, intellectual properties~\cite{wang2024did,wang2024trace}, or creating disturbing and improper subject matter, including eroticism and violence~\cite{schramowski2023safe}. 
Addressing these issues necessitates continuously removing those improper concepts from these models to prevent misuse and protect copyright from infringement.

Existing techniques aiming to remove concepts from the text-to-image diffusion models can be categorized into two types. For a given concept that needs to be removed, the first group of methods refines the training data by discarding images containing the undesired concept and then retrains the model from scratch~\cite{nichol2021glide, openai, schuhmann2022laion}. The other set of methods removes the target concept without requiring full retraining.
These methods instead utilize a small amount of additional data to fine-tune the models and modify specific neurons~\cite{2024-UCE, 2023-ESD,dai2021knowledge}.
In real-world scenario, the improper concepts learned by the models such as copyright-protected art styles often discovered by the model owner in a continuous manner.
For example, various artists may continually raise complaints that text-to-image generative AI can replicate their distinctive art style.
Additionally, users or red-teaming teams~\cite{chin2023prompting4debugging} of these models may continuously flag instances where the models generate harmful or malicious contents.
However, we find that these existing techniques do not perform well in scenarios where different concepts need to be continuously removed one after another, which is a practical and important use case.
In detail, we observe that training data filtering methods require model owners to retrain the model from scratch, which is deemed impractical due to its exorbitant cost.
The fine-tuning-based methods often struggle to maintain the alignment between the text prompts and the generated images after repeated removals, degrading the quality and coherence of the generated content (we discuss such ``entity forgetting'' problem in \autoref{sec:motivation}).
Thus, it is important to design a method that can continuously remove improper concepts learned by the text-to-image diffusion models with low costs.

In this paper, we propose an approach called \ours{}(\textbf{C}ontinuous \textbf{C}oncepts \textbf{R}emoval in \textbf{T}ext-to-image
Diffusion Models) to remove concepts continuously while keeping the text-image alignment of the model. 
Specifically, we develop a knowledge distillation paradigm that concurrently eliminates the unwanted concepts from the model while ensuring the edited model's generation quality and text prompt comprehension ability remain aligned with the original model. This is accomplished by utilizing a collection of prompts produced through our genetic algorithm, which incorporates a designed fuzzing strategy.
Through extensive experiments, we demonstrate the effectiveness of \ours{} in removing a variety of concepts continuously. Our results, evaluated using both automated metrics and human studies, show that \ours{} can effectively excise targeted concepts such as specific artistic styles and improper content while preserving the text-image alignment of the model, ensuring that the output remains faithful to the intended textual descriptions. For example, while keeping continuous concept removal at an average removal rate of 0.87, our method improves the CLIP score from 21.698 to 25.005 compared to the existing state-of-the-art. 

Our contributions are summarized as follows:
\circled{1} We introduce the continuous concept removal problem, which better represents real-world situations and has more practical applications.
\circled{2} We find that existing methods do not work well in the continuous concept removal. In detail, we find these methods lead to poor
alignment between the text prompts and the generated image after the continuous
removal process.
\circled{3} We propose a novel approach \ours{} that can effectively remove concepts continuously while keeping text-image alignment of the text-to-image diffusion models.
\circled{4} We conduct a comprehensive evaluation, including automated evaluation and human study. Our experimental results demonstrate \ours{} significantly outperforms the state-of-the-art concept removal methods in the continuous concept removal problem.

\vspace{-0.1cm}
\section{Related Work}
\label{sec:back_related_work}
\vspace{-0.1cm}

\noindent\textbf{T2I diffusion models.}
Text-to-image (T2I) diffusion models have made significant progress in image synthesis tasks~\cite{reed2016generative,li2019controllable,yu2022scaling,ramesh2021zero}.
One of the most notable open-sourced text-to-image diffusion models is Stable Diffusion~\cite{rombach2022high}. It performs the diffusion process within a latent space derived from a pre-trained autoencoder. The autoencoder reduces the dimensionality of the data samples. Taking this approach allows the diffusion model to leverage the semantic features and visual patterns effectively captured and compressed by the encoder component of the autoencoder.

\noindent\textbf{Concept removal on T2I diffusion models.}
With the advancements of text-to-image diffusion models, there are also many misuse problems surrounding around them~\cite{setty2023ai,schramowski2023safe,struppek2022biased,bansal2022well}.
The generated content of the text-to-image diffusion models can infringe established artistic styles~\cite{ghosh2022can} or contain improper concepts like pornography and violence~\cite{schramowski2023safe}.
Concept removal is a promising way to defend against the misuse problems of diffusion models~\cite{kumari2023ablating,2023-ESD,2024-UCE}. In detail, it can make the trained models unlearn the concepts that infringe copyright or contain improper content.
The concept removal in the text-to-image diffusion models can be view ``model editing'' process~\cite{2023-ESD, 2024-UCE, karras2020training, kumari2023ablating, liu2020towards, zhang2023forget, ojha2021few, li2024towards} achieved by fine-tuning/modifying the model weights. Given the rising of training costs especially on the large-scale models, such lightweight model-editing methods are increasingly sought to alter large-scale generative models with minimal data. 
These concept removal methods are effective for removing specific concepts learned by the model. However, we find that these existing methods do not perform well in the scenario where the concepts need to be continously removed.

\vspace{-0.1cm}
\section{Motivation}
\label{sec:motivation}
\vspace{-0.1cm}

In this section, we introduce the motivation for our approach. 
We begin by highlighting the importance of continuously removing concepts. We then demonstrate that existing techniques fail to remove the concepts continuously while keeping high generation quality of the model.

\vspace{-0.1cm}
\subsection{Necessity of continuous concept removal}
\label{subsec:necessity_continuously_removing_concepts}
\vspace{-0.1cm}

With the development of text-to-image diffusion models, there is a growing need to prevent their misuse, such as generating malicious content or infringing on copyrights. Removing certain concepts from these models shows promise in addressing this issue. However, model owners/governors often continuously discover improper concepts (e.g., those involving violence or specific artists' copyrighted styles) that the models have learned. For instance, different artists may continuously claim that text-to-image diffusion models like DALL-E 3~\cite{betker2023improving} and Midjourney~\cite{midjourney} can mimic their distinctive styles.
Additionally, users may continuously report the generation of malicious content such as violence, guns, and nudity by these models.
Thus, model owners/governors require a technique that can \emph{swiftly and continuously remove the improper concepts} from the deployed models.

\vspace{-0.1cm}
\subsection{Limitation of existing techniques}
\label{subsec:existing_limitation}
\vspace{-0.1cm}

\begin{figure*}[!htb]
    \centering
    \includegraphics[width=0.9\linewidth]{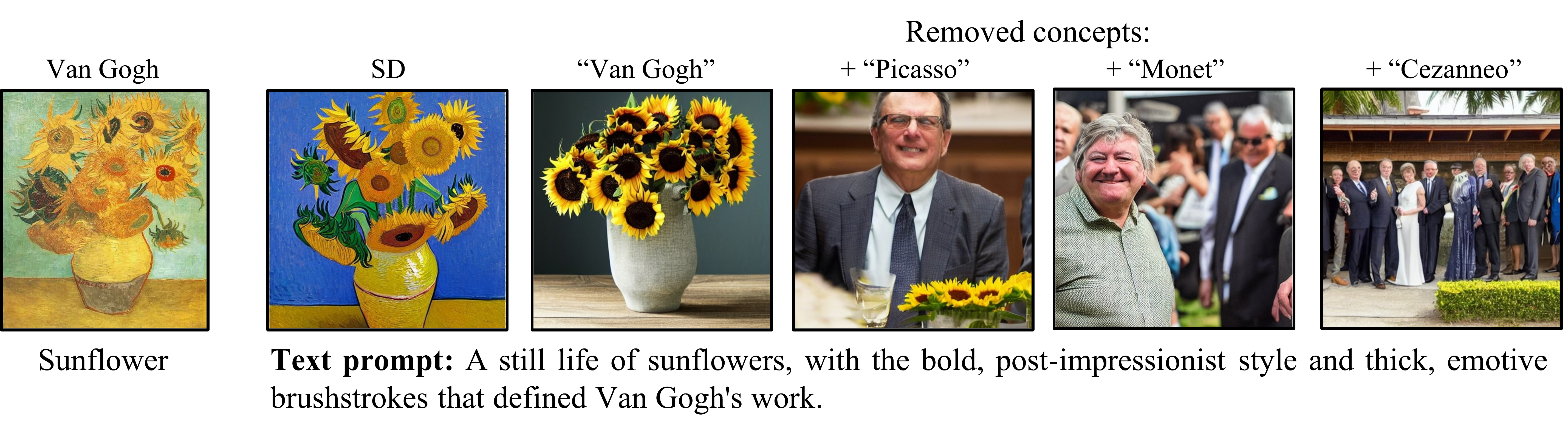}
    \caption{The performance of ESD on removing concepts continuously. It showcases the progress of ESD continuously removing concepts and the generated images of a fixed text prompt.
    The leftmost is a true art work of Van Gogh. The right images are generated by Stable Diffusion (SD), ESD (removing ``Van Gogh''), ESD (removing ``Van Gogh'' +``Picasso''), ESD (removing ``Van Gogh'' +``Picasso'' + ``Monet''), and ESD (removing ``Van Gogh'' +``Picasso'' + ``Monet'' + ``Cezanneo''),  respectively.
    Observe that the text-image alignment is continuously destroyed as the concept removal process continues.}
    \label{fig:motivation1}
    \vspace{-8pt}
\end{figure*}
A straightforward solution to the issue of continuous concept removal is to reemploy existing techniques whenever a new concept requires removal.
Among them, ESD~\cite{2023-ESD} is the most representative, which formalizes concept removal into optimization to eliminate the influence of concept $\vx$.
However, a problem arises during optimization as concepts are not isolated but interconnected with other related concepts. This means that when ESD attempts to eliminate a specific concept $\vx$, it causes a shift in the semantic space of diffusion models.
For instance, removing the concepts of artists continuously, such as  ``Van Gogh'', ``Picasso'', ``Monet'' and ``Cezanneo'', also affects the concept of ``sunflowers''.
Such an incidental semantic space shifting becomes more serious as the concept removal continues.
~\autoref{fig:motivation1} illustrates the problem visually.
\begin{figure}[!t]
    \centering
    \includegraphics[width=0.49\textwidth]{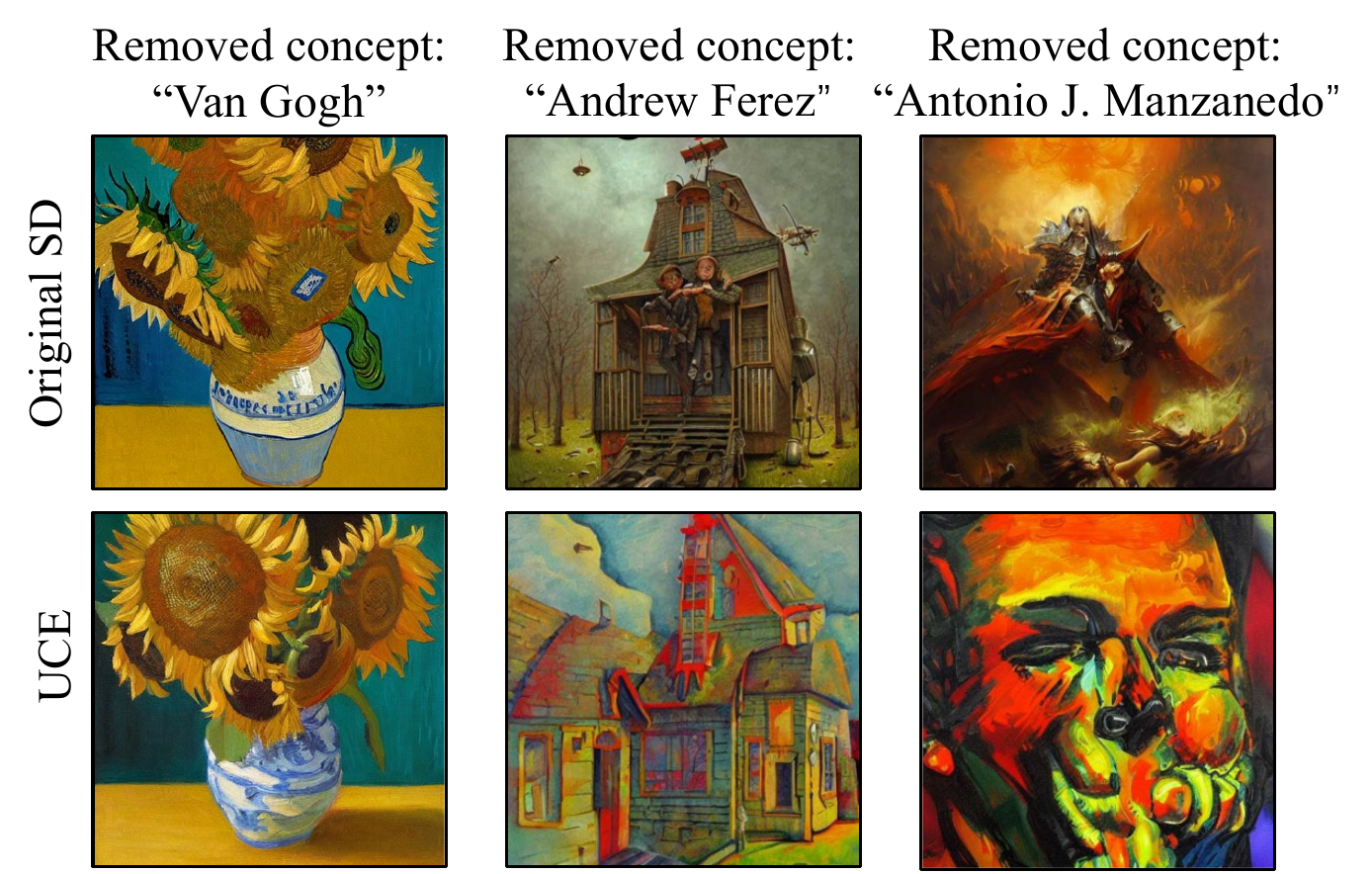}
    \caption{Performance of UCE when removing different concepts. Observe that UCE succeeds in removing the concepts “Andrew Ferez” and “Antonio J. Manzanedo” but fails for “Van Gogh”.}
    \label{fig:motivation_UCE}
    \vspace{-12pt}
\end{figure}

We present a comparison between the original work of Sunflowers by ``Van Gogh'' and images produced by diffusion models edited by ESD when various artistic styles are progressively deleted. 
The process involves iteratively removing styles, starting with ``Van Gogh'', and then proceeding to eliminate additional styles, ``Picasso''.
The outcomes displayed in ~\autoref{fig:motivation1} reveal a concerning trend where the alignment between the text prompt and the generated image deteriorates with each incremental removal of artistic styles. 

Initially, when removing the ``Van Gogh'' style, the image closely corresponds to the text prompt.
However, as the removal continues, the images deviate further from the original prompt. 
When removing the ``Picasso'' style, the focus shifts towards another concept, leading to images primarily featuring portraits with sunflowers playing a minor role. 
This progression highlights how the iterative use of ESD results in significant shifts within the semantic space. 
We define this observation as \textbf{entity forgetting}, where the models are challenged to maintain a coherent understanding of entities such as ``sunflowers'' over continuous iterations of concept removal. 
There are also techniques like UCE~\cite{2024-UCE} aiming to remove multiple concepts simultaneously.
However, in practical application, we find that their performance fluctuates greatly across different concepts.
\autoref{fig:motivation_UCE} showcases some examples.
Observe that UCE shows satisfactory performance on certain concepts.
However, regarding specific concepts like ``Van Gogh'', UCE does not deliver the same effectiveness.
Due to extensive and relevant training data associated with concepts like ``Van Gogh'', such concepts are deeply embedded in the model representations and difficult to remove entirely.

\vspace{-0.1cm}
\section{Method}
\label{sec:method}
\vspace{-0.1cm}
To remove concepts continuously and avoid \textit{entity forgetting}, we propose \ours{}.
Our approach relies on the knowledge distillation paradigm, which simultaneously removes concepts (removing unwanted knowledge) and aligns the latent semantic space between the original Stable Diffusion models and the edited ones (preserving essential knowledge for text-image alignment).
Besides the loss designed for concepts removal, \ours{} also incorporates a regularization loss to align the semantic space, whenever a new concept is required to be removed.
Additionally, \ours{} features an entity generation mechanism combining genetic algorithm and fuzzing strategy to generate the searched calibration prompt used in the regularization to enhance effectiveness.

\vspace{-0.1cm}
\subsection{Distillation for concepts removal and alignment}
\label{subsec:conc_removal}
\vspace{-0.1cm}

\textbf{Problem formulation.}
The primary objectives of \ours{} are removing concepts continuously and keeping the text-image alignment.
The removal target can be formulated as:
\vspace{-0.1cm}
\begin{equation}
    \label{eq:removal_target}
    \epsilon_\theta(\vx_\vt, \vt) \gets \epsilon_\theta(\vx_\vt, \vc, \vt), \quad \forall \vc \in \mathcal{C} 
\end{equation}
where $\vx_\vt$ represents the image $\vx$ stamped by a noise at timestep $\vt$, $\mathcal{C}$ the latent concept set to be eliminated, and $\epsilon_{\theta}$ the diffusion model under concept removal. Intuitively, \autoref{eq:removal_target} indicates making $\epsilon_\theta(\cdot)$ ignore the influence of concept $\vc$.
On the other hand, the target to keep the text-image alignment can be formulated as:
\vspace{-0.1cm}
\begin{equation}
    \label{eq:alignment_target}
    \epsilon_{\theta^*}(\vx_\vt, \vp, \vt) \gets \epsilon_\theta(\vx_\vt, \vp, \vt), \vp \in \mathcal{P} \backslash \mathcal{C}
\end{equation}
where $\epsilon_{\theta^*}$ denotes the original diffusion model with frozen parameters and $\mathcal{P} \backslash \mathcal{C}$ the input prompt space $\mathcal{P}$ that doesn't contain concept $\mathcal{C}$.
\autoref{eq:alignment_target} indicates that \ours{} should keep the alignment as the stable diffusion model when given text prompts that are irrelevant to removed concepts.

\noindent
\textbf{Continuous concept removal.}
Given the original diffusion model with frozen parameters $\epsilon_{\theta^*}(\cdot)$, we aim to remove concept $\vc$ on the diffusion model $\epsilon_\theta(\cdot)$ initialized by $\epsilon_{\theta^*}(\cdot)$.
Following the previous work~\cite{2023-ESD}, we quantify the negative removal guidance direction of $\vc$ as follows:
\begin{equation*}
\Delta_\vc = 
    \epsilon_{\theta^*}\left(\vx_{\vt}, \vt\right)-\eta\left[\epsilon_{\theta^*}\left(\vx_{\vt}, \vc, \vt\right)-\epsilon_{\theta^*}\left(\vx_{\vt}, \vt\right)\right]
\end{equation*}
In particular, the term $\left[\epsilon_{\theta^*}\left(\vx_{\vt}, \vc, \vt\right)-\epsilon_{\theta^*}\left(\vx_{\vt}, \vt\right)\right]$ represents the additional impact of concept $\vc$ on noise prediction.
The removal loss is adapted from it and deployed iteratively as follows:

\vspace{-0.2cm}
\begin{equation}  
    \label{eq:removal_loss}
    \mathcal{L}_{rm}= \Vert\epsilon_\theta(\vx_\vt, \vc, \vt) - \Delta_\vc \Vert_p
\end{equation}
where $\vx_\vt$ denotes the generated images and $\vt$ the step of noise in diffusion process.
$\Vert \cdot \Vert_p$ is the $p$ norm ($p=1$ in our paper).
The \autoref{eq:removal_loss} aims to guide $\epsilon_\theta(\vx_\vt, \vc, \vt)$ to a direction that contains no effect of $\vc$.
Note that \autoref{eq:removal_loss} operates iteratively and follows a memoryless property, meaning that each iteration builds on the model from the previous step rather than the original diffusion model.
This approach enables \ours{} to adapt to new requirements as they emerge dynamically.

\noindent
\textbf{Text-image alignment.}
As shown in \autoref{sec:motivation}, iteratively deploying single \autoref{eq:removal_loss} results in a serious \emph{entity forgetting}, disrupting text-image alignment severely.
Subsequently, we introduce an alignment loss to regulate the model's behavior.
With some generated entity-related text prompts (called calibration prompt set), we deploy the alignment regularization loss:
\begin{equation}
    \label{eq:reg_loss}
    \mathcal{L}_{reg} = MSE(\epsilon_\theta(\vx_\vt, \ve, \vt), \epsilon_{\theta^*}(\vx_\vt, \ve, \vt)), \quad \ve \in \mathcal{E}
\end{equation}
where $\mathcal{E}$ and $\epsilon_\theta(\vx_\vt, \ve, \vt)$ denote the calibration prompt set and the noise prediction of an entity text prompt $\ve$ (e.g., ``a picture of sunflower''), respectively.
$MSE(\cdot)$ denotes the mean square error function~\cite{allen1971mean}.
Model $\epsilon_{\theta^*}(\cdot)$ means the original diffusion model with frozen parameters, which is taken as the teacher net.
The $\mathcal{L}_{reg}$ is designed to regularize $\epsilon_\theta$($\cdot$), the student net, to mimic the teacher's behavior, $\epsilon_{\theta^*}(\cdot)$.
$\mathcal{L}_{reg}$ enables student net $\epsilon_\theta(\cdot)$ to approximate teacher net $\epsilon_{\theta^*}(\cdot)$'s entity understanding ability, overcoming ``entity forgetting''.

\noindent
\textbf{Knowledge distillation paradigm.}
A knowledge distillation paradigm is employed to achieve continuous concept removal and maintain text-image alignment simultaneously.
We formulate it into an optimization problem with the definitions of $\mathcal{L}_{rm}$ and $\mathcal{L}_{reg}$:
\begin{equation}
    \label{eq:dis_loss}
     \min_{\epsilon_\theta} \mathcal{L} = \mathcal{L}_{rm} + \lambda \cdot \mathcal{L}_{reg}
\end{equation}
where $\lambda$ is the hyper-parameter to balance $\mathcal{L}_{rm}$ and $\mathcal{L}_{reg}$, $\lambda \geq 0$.
\ours{} addresses the task of continuous concept removal with text-image alignment by optimizing $\mathcal{L}$ via gradient descent, yielding an ideal edited model, $\epsilon_\theta$($\cdot$).
During the continuous removal process, assume we want to remove concept $\vc_i$ at the removal step $i$.
Given the original diffusion model $\epsilon_{\theta^*}(\cdot)$ and the model from previous step $\epsilon_\theta^{i-1}(\cdot)$, which has removed concepts $\{\vc_1, \vc_2, ..., \vc_{i-1}\}$, we obtain $\epsilon_\theta^{i}(\cdot)$ by deploying distillation between $\epsilon_{\theta^*}(\cdot)$ and $\epsilon_\theta^{i-1}(\cdot)$ on concept $\vc_i$ through loss $\mathcal{L}$, defined in~\autoref{eq:dis_loss}.

\begin{figure}[t]
    \centering
    \includegraphics[width=0.99\linewidth]{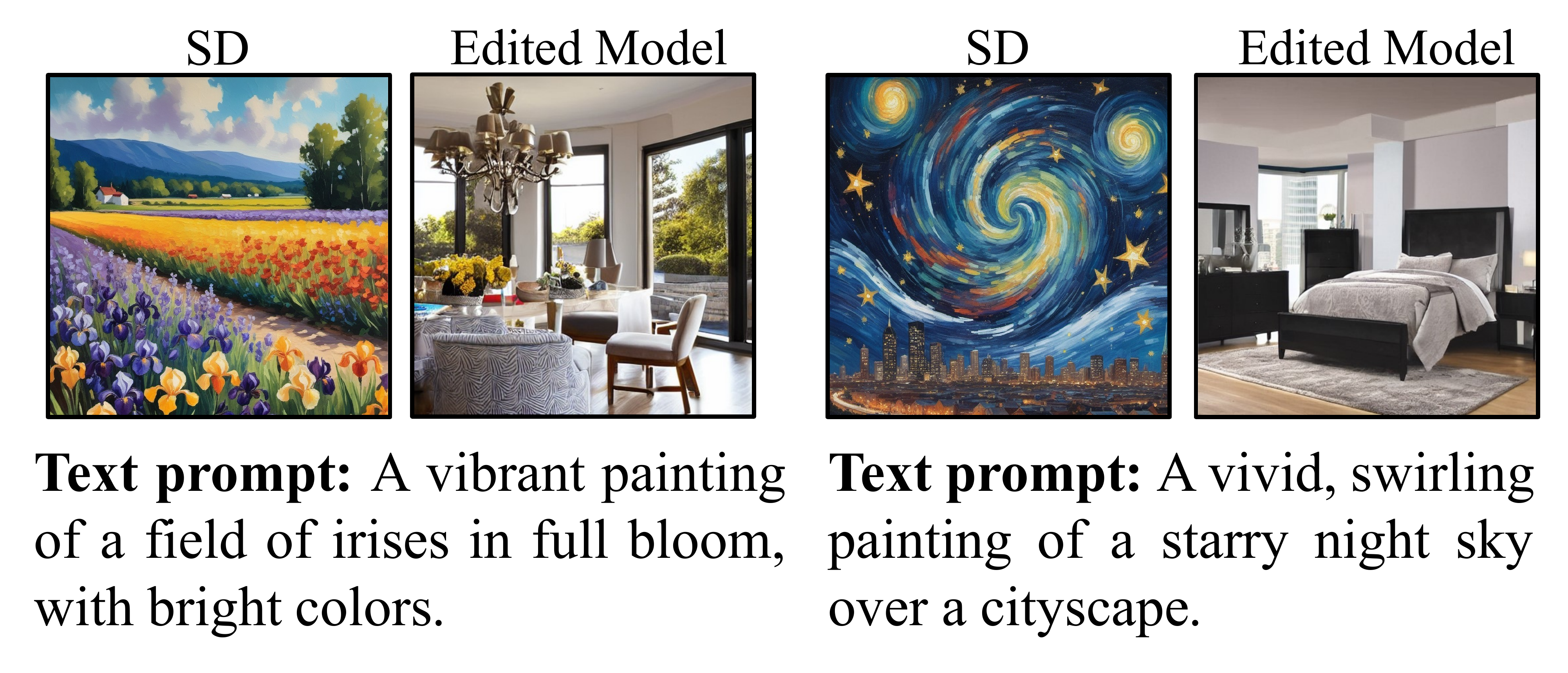}
    \vspace{-0.2cm}
    \caption{Performance of distillation with text prompts on random entities. For each example, the left one is generated by edit models and the right one by the original T2I diffusion model (T2IDM). Observe that text-image alignment is terrible in some cases.}
    \label{fig:perfor_random_entity}
    \vspace{-0.5cm}
\end{figure}

\noindent
\textbf{Necessity of optimized calibration prompt set.}
During our practice of \autoref{eq:dis_loss}, we find that a random calibration prompt set causes an unstable text-image alignment.
\autoref{fig:perfor_random_entity} supports the evidence.
We deploy \autoref{eq:dis_loss} with a calibration set based on randomly selected entities.
The left image is generated by the original diffusion model (Stable Diffusion specifically) of the corresponding text prompt, and edited models generate the right one. 
The results exhibit oscillatory and unstable behavior, indicating that existing methods perform well in some cases but poorly in others.
Specifically, the distillation can maintain text-image alignment for some entities but may have misalignment for others. 
This variability arises because distillation  different entities used in distillation impacts semantic matching to varying degrees.
In some specific entities, the misalignment becomes more severe (the right case of \autoref{fig:perfor_random_entity}), and we need to harden such semantics.
A higher misalignment between the entities and generated images indicates semantics that are more vulnerable and, therefore, prioritized for hardening.
With only a calibration prompt set based on some randomly generated entities, distillation aligns only random parts of the semantic space, leading to the undesired results shown in \autoref{fig:perfor_random_entity}.
To address this, we optimize the calibration prompt set to generate entities needing alignment most.
Such entities serve as anchors within the semantic space, and the entire space is aligned through these entities.

\begin{figure}
    \centering
    \begin{minipage}[t]{0.46\textwidth}
        \centering
        \includegraphics[width=\textwidth]{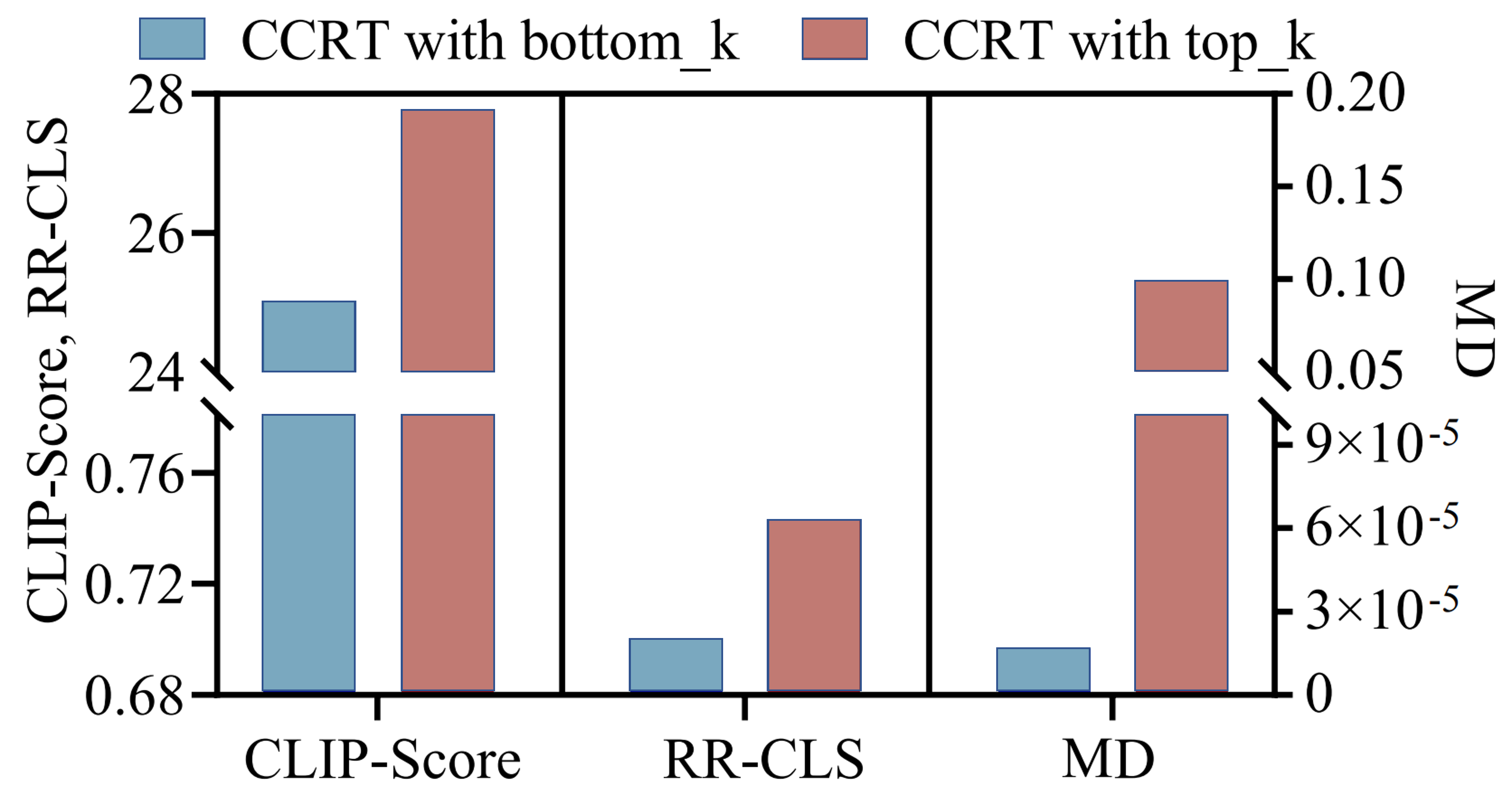}
    \vspace{-18pt}
    \caption{Performance of \ours{} with entities having top k and bottom k \emph{Misalignment Distance (MD)} values. CLIP-Score $\uparrow$. RR-CLS $\uparrow$. Note that a higher MD value is associated with increased CLIP-Score and RSR-CLS.}
    \label{fig:meth_motivation_1}
    \end{minipage}
    \vspace{-12pt}
\end{figure}

\subsection{Calibration prompt set generation}
\label{subsec:eneity_generation}
In this section, we introduce how to generate the calibration prompt set to improve distillation performance.
Recent research~\cite{shrivastava2016training, takac2013mini} demonstrates that hard sample mining improves model performance to a great extent.
Hard sample mining targets to identify the samples that are most useful to a specific task.
For text-image alignment, hard samples refer to entities of which the model's semantic space is more broken, resulting in a more severe misalignment.
Inspired by it, we introduce a calibration prompt set generation mechanism to mine hard entities and improve distillation performance.
Following the definition of $\mathcal{L}_{reg}$ in \autoref{eq:alignment_target}, we propose \emph{Misalignment Distance (MD)} based on norm to measure hardness and identify hard entities:
\vspace{-0.2cm}
\begin{equation}
    \label{eq:distance}
    MD (\epsilon_{\theta^{*}}, \epsilon_\theta, \ve_i) = \dfrac{1}{N}\sum_{i=1}^N{||\epsilon_\theta(\ve_i)-\epsilon_{\theta^{*}}(\ve_i)||_p}, \quad \ve_i \in \mathcal{E}
\end{equation}
\vspace{-0.3cm}

where $\mathcal{E}$ denotes the calibration set, initialized by entities from ImageNet classes~\cite{russakovsky2015imagenet}. $\epsilon_\theta(\cdot)$ means the diffusion model to be removed concepts and $\epsilon_{\theta^{*}}(\cdot)$ the original diffusion model.
The higher MD indicates the more misalignment, the more important we need to reinforce corresponding semantics.
Note that at this step, the calibration set consists of entities without accompanying prompt texts.
We sort $\mathcal{E}$ by \autoref{eq:distance} and select the top $k$ entities ($k=10$).
To validate the impact difference between entities, we also select the bottom $k$ entities as a control group.
Two metrics, \emph{RR-CLS} and \emph{CLIP-Score}, are utilized to evaluate the concept removal ability and text-image alignment, respectively.
A higher CLIP-Score means a better text-image alignment, while a higher RR-CLS reflects a better concept removal ability.
Details of the definition can be found in 
(\autoref{subsec:experiment_setup}). 
\autoref{fig:meth_motivation_1} presents the results, where the red bar is taller than the blue one on CLIP-Score and RR-CLS, indicating that ``\textit{CCRT with top k}'' performs better than ``\textit{CCRT with bottom k}''.
Considering MD, we conclude that entities with higher MD result in better distillation performance.
To mine such hard entities, heuristic algorithms (genetic algorithm specifically) are considered.
The genetic algorithm is well-suited for complex problems such as hard entity mining, as it efficiently explores large search spaces and evolves solutions to identify valuable entities~\cite{alhijawi2024genetic,holland1992genetic}.
To expand the diversity of found entities, we embed a fuzzing strategy enhanced by large language model (LLM), which will generate more diverse entities through specific rules.
The terminologies used are summarized in \autoref{tab:notion_overview}.

\begin{figure*}[thb]
    \centering
    \includegraphics[width=0.93\textwidth]{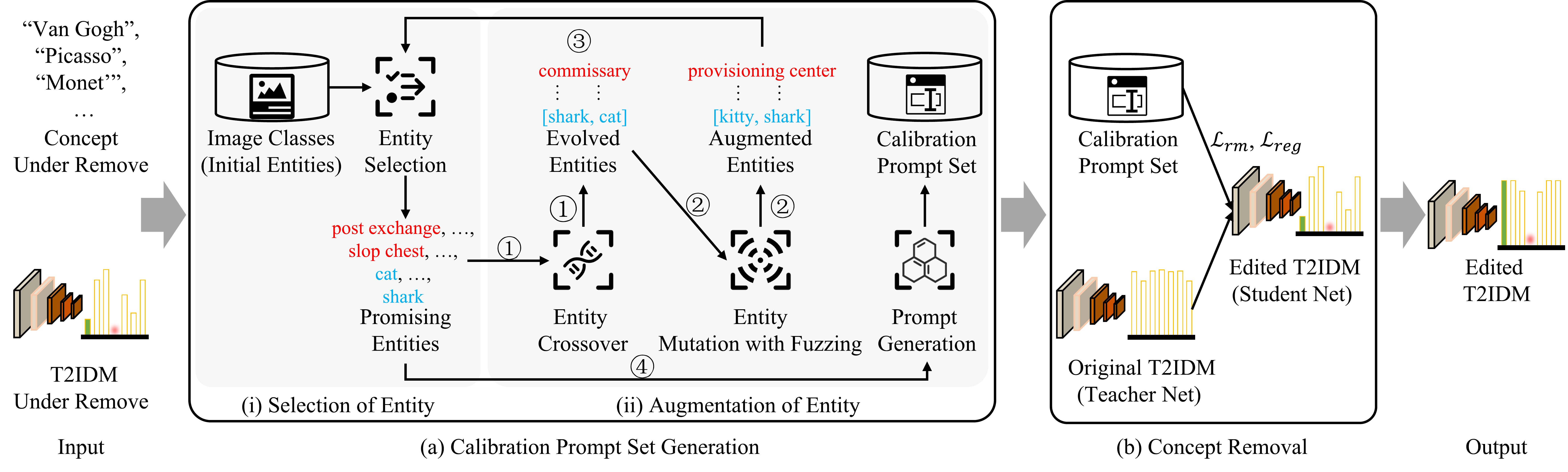}
    \vspace{-0.3cm}
    \caption{The overview of \ours{} on text-to-image diffusion model (T2IDM). \ours{} divides the continuous concept removal task into two stages: calibration prompt set generation (a) and concept removal (b). In the first stage, \ours{} utilizes a genetic algorithm to generate prompts as a calibration set (a). Subsequently, \ours{} utilize the calibration set to remove concepts with a distillation mechanism (b).}
    \label{fig:framework}
    \vspace{-12pt}
\end{figure*}

\begin{algorithm}[!t]
\small
  \caption{Genetic Algorithm with Fuzzing}
  \label{alg:genetic_algorithm}
  \begin{algorithmic}[1] 
    \REQUIRE Initialized Entity Set: $\mathcal{E}$, Optimization Direction: $MD$, Original and Edited Diffusion Models: $\epsilon_{\theta^*}$, $\epsilon_{\theta}$, Generation Threshold: $G$
    \ENSURE Calibration set
    \STATE $\mathcal{E} \gets \mathcal{E}$ sorted by $MD(\epsilon_{\theta^*}, \epsilon_{\theta}, \mathcal{E})$ 
    \STATE $\mathcal{E} \gets$ Top-k$(\mathcal{E})$, $\mathcal{E'} \gets \emptyset$, $\vg \gets 1$
    \REPEAT
    \STATE $pars \gets$   $select\_pars(\mathcal{E})$, $\mathcal{E'} \gets pars$
    \FOR{i = 1, 3, ..., $\lfloor len(pars) / 2 \rfloor$}
        \STATE $par_1 \gets pars[i]$,  $par_2 \gets pars[i+1]$
        \STATE $child \gets crossover(par_1, par_2)$
        \STATE $child \gets mutaion\_fuzzing(child)$
        \STATE $\mathcal{E'}, g \gets  \mathcal{E'} \cup child, g+1$
    \ENDFOR
    \STATE $\mathcal{E} \gets$ Top-k$(\mathcal{E} \cup \mathcal{E'})$
    \UNTIL $\vg \leq G$  
    \RETURN $\mathcal{E}$
  \end{algorithmic}
\end{algorithm}

To further explore potential entities with more hardness, we propose Algorithm~\ref{alg:genetic_algorithm}, featuring a genetic algorithm with a fuzzing strategy enhanced by LLM.
We first initialize the calibration set by image classes from ImageNet, with each \emph{individual} containing one entity to start.
An individual means an element of the calibration set, consisting of a list of entities, for example, [\textit{``post exchange''}].
Algorithm~\ref{alg:genetic_algorithm} aims to optimize the calibration set towards increased environment fitness. 
The optimization direction of each element is evaluated through~\autoref{eq:distance}. 
The terminologies and their meaning are summarized in \autoref{tab:notion_overview}.
\autoref{fig:meth_motivation_1} shows that higher MD values identify entities with greater potential to enhance distillation performance. 
We first sort the initialized entity set by MD and select the top-k entities (lines 1-2).
Then, we randomly select individuals as \emph{parents} (i.e., the individuals used to generate new ones) from $\mathcal{E}$ and assign them to a temporary list, $pars$, with $\mathcal{E}'$ updated to include the selected individuals (lines 4-6).
To generate new high-quality entities with increased MD, we introduce \texttt{crossover} for optimization (line 7).
\texttt{crossover} combines two individuals to create a new one by two specified rules.
On the one hand, if entities of the individuals have a shared parent, the generated individual will be the parent entity. 
The semantic hierarchy of ImageNet classes is referred to \cite{ImageNetHierarchy}.
For example, the individual generated from the parent individuals [``\textit{post exchange}''] and [``\textit{slop chest}''] is [``\textit{commissary}''], reflecting their semantic relation.
On the other hand, if there is no semantic relationship between the entities, the generated individual will combine both parent entities.
For instance, [``\textit{cat}'', ``\textit{shark}''] is generated from [``\textit{cat}''] and [``\textit{shark}''].
However, \texttt{crossover} is limited to identifying entities within the initial ImageNet image classes. 
To discover high-quality entities with greater MD from a broader search space, we introduce a strategy called \texttt{mutation\_fuzzing} to generate additional, similar high-quality entities, where \ours{} employs LLM to replace randomly selected entities with synonyms (line 8).
The \texttt{mutation\_fuzzing} can be divided into two stages, \texttt{mutation} and \texttt{fuzzing}.
The \texttt{mutation} replaces randomly selected entities with synonyms identified by LLM, specifically GPT-4 in our implementation.
For example, the result of \texttt{mutation}([\textit{``cat'', ``shark''}]) might be [\textit{``kitty'', ``shark''}], where \textit{``cat''} is replaced with \textit{``kitty''}.
We then implement a fuzzing strategy to generate additional entities based on the initial set. 
The \texttt{fuzzing} leverages LLM to create large batches of data, expanding the calibration set with potentially high-quality entities (detailed in \autoref{subsec:crossovber_mutation}). 
For example, \texttt{fuzzing}([\textit{``coffee mug''}]) might produce [\textit{``desk lamp'', ``backpack'', ``pencil case''}].
More details about \texttt{crossover} and \texttt{mutation\_fuzzing} can be found in \autoref{subsec:crossovber_mutation}.
We then add these generated entities to $\mathcal{E}'$ and repeat the generation iteratively until it reaches a threshold pre-defined by the developer (lines 9-12).

With generated high-quality entities, \ours{} then uses LLM to combine entities into semantically coherent text prompts to craft the final calibration prompt set.
For instance, an individual with entities [\textit{``snowbird'', ``kitty''}] might be combined into the text prompt: ``A vibrant snowbird perched next to a colorful kitty in a lush tropical setting.''
The prompt for LLM can be found in \autoref{subsec:data_overview}. 
Such text prompts consist of the \textbf{calibration prompt set} used in distillation to ensure text-image alignment.

\subsection{Procedure of \ours{}}
\label{subsec:overview}
\autoref{fig:framework} illustrates the continuous concept removal procedure of \ours{}.
It consists of two main components: (a) calibration prompt set generation and (b) concept removal.

Given the original diffusion model and specific concept under removal, \ours{} first utilizes the entities from ImageNet (image classes) as the initial set.
Then, we employ an elaborate hardness identification function, defined by~\autoref{eq:distance}, ensuring the selection of the most promising entities for the next phase.
After selection, \ours{} uses \texttt{crossover}
to evolve the calibration set (\circled{1}) and \texttt{mutation\_fuzzing}
to expand the calibration set (\circled{2}).
The key intuition behind \texttt{crossover} and \texttt{mutation\_fuzzing} is to construct entities with higher MD that can act as better semantic anchors, thereby stabilizing the text-image alignment of the edited models. 
With entities from \texttt{crossover} and \texttt{mutation\_fuzzing}, \ours{} selects the most promising candidates according to \autoref{eq:distance} (\circled{3}).
\ours{} then iteratively applies \texttt{crossover} and \texttt{mutation\_fuzzing} to these refined entities (\circled{4}) until it reaches a threshold predefined by the developer.
\ours{} then feeds the entity set into LLM to weave semantically coherent text prompts for each individual (\circled{4}), finally outputting the calibration prompt set.
In phase (b), a distillation process is implemented. The original diffusion model serves as the teacher net to keep the text-image alignment of the edited model, while the student net is edited to remove specific concepts such as ``Van Gogh''. 
This modification ensures concept removal and text-image alignment simultaneously, which is achieved through the generated calibration prompt set.

\section{Experiments}
\label{sec:experiments}
We apply our proposed method, called \ours{}, to the widely employed diffusion model known as Stable Diffusion (SD v1.4 by default)~\cite{rombach2022high}.
Our experimental evaluation comprises two distinct components: an automated evaluation and a user study, in which human participants conduct assessments and judgments.
We evaluate \ours{} by answering the following research questions(RQs):
\begin{itemize}
    \item RQ1. How effective is \ours{} in continuous concept removal, such as artist style, improper content, and Intellectual Property (IP)?
    \item RQ2. How effective is \ours{} in text-image alignment?
    \item RQ3. Ablation study of \ours{}?
    \item RQ4. How efficient is \ours{}?
\end{itemize}
Due to space limitations, we put the results for RQ3 and RQ4 in the Appendix.%

\subsection{Experiment Setup}
\label{subsec:experiment_setup}
\begin{table}[!b]
    \vspace{-15pt}
    \centering
    \footnotesize
    \tabcolsep=5pt
    \caption{Results of \ours{} on continuous improper content removal. RR-CLS is taken as the metric. RR-CLS $\uparrow$.}
    \vspace{-0.2cm}
    \label{tab:improper_content}
    \begin{tabular}{cccc}
        \toprule
            \multirow{2}{*}{Improper Concent} & \multirow{2}{*}{SD} & \multicolumn{2}{c}{\ours{} (Ours)} \\
            \cmidrule(lr){3-4}
             &  & ``Eroticism'' & + ``Violence'' \\
             \cmidrule(lr){1-4}
            ``Eroticism'' & 0.39 & 0.95 & 0.99 \\
            ``Violence'' & 0.51 & 0.69 & 0.93 \\
            \bottomrule
    \end{tabular}
\end{table}

\begin{table*}[th]
    \centering
    \footnotesize
    \tabcolsep=5pt
    \caption{Comparison of \ours{} and other techniques on the effectiveness for continuous artistic style removal.
    Four famous artistic styles are removed continuously in the order of ``Van Gogh'', ``Picasso'', ``Monet'', ``Cezanne''.
    Observe that \ours{} achieves 0.753 and 0.874 on RR-CLS and RR-LLM on average.
    RR-CLS $\uparrow$, RR-LLM $\uparrow$.}
    \vspace{-0.2cm}
   \begin{tabular}{lcccccccccc}
   \toprule
    \multirow{2}{*}{\begin{tabular}[c]{@{}c@{}}Removed\\ Concept\end{tabular}} & \multicolumn{2}{c}{SD} & \multicolumn{2}{c}{UCE} & \multicolumn{2}{c}{MACE} & \multicolumn{2}{c}{SPM} & \multicolumn{2}{c}{\ours{} (Ours)} \\
    \cmidrule(lr){2-3} \cmidrule(lr){4-5} \cmidrule(lr){6-7} \cmidrule(lr){8-9} \cmidrule(lr){10-11} 
     & RR-CLS & RR-LLM & RR-CLS & RR-LLM & RR-CLS & RR-LLM & RR-CLS & RR-LLM & RR-CLS & RR-LLM \\
     \midrule
    ``Van Gogh'' & 0.150 & 0.014 & 0.393  & 0.071  & 0.471  & 0.271  & 0.386  & 0.286  & \textbf{0.743 } & \textbf{0.757 } \\
    +``Picasso'' & 0.000 & 0.055 & 0.124  & 0.008 & 0.376  & 0.104  & 0.224  & 0.072  & \textbf{0.712 } & \textbf{0.872 } \\
    +``Monet'' & 0.140 & 0.160 & 0.100  & 0.060  & 0.353  & 0.147  & 0.233 & 0.100  & \textbf{0.740} & \textbf{0.947 } \\
    +``Cezanne'' & 0.186 & 0.013 & 0.241  & 0.044  & 0.423  & 0.077  & 0.373  & 0.159  & \textbf{0.818 } & \textbf{0.918 } \\
    Average & 0.119 & 0.061 & 0.215  & 0.046  & 0.406 & 0.150  & 0.304  & 0.154  & \textbf{0.753 } & \textbf{0.874 } \\
    \bottomrule
    \end{tabular}
    \label{tab:main_eval}
    \vspace{-0.1cm}
\end{table*}

\begin{table*}[th]
    \centering
    \begin{minipage}[t]{0.74\textwidth}
        \centering
        \footnotesize
        \tabcolsep=4pt
      \caption{Results of the human evaluation. Detailed instructions are provided in~\autoref{subsec:human_evaluation_instruction}.
      The values represent the average rank assigned to each method for a given target concept. A higher rank (closer to 1) indicates better performance on the corresponding dimension.}
      \vspace{-0.2cm}
        \label{tab:human_study}
       \begin{tabular}{lcccccccccccc}
       \toprule
        \multirow{2}{*}{Target Concept} & \multicolumn{3}{c}{Concept Removal} & \multicolumn{3}{c}{Text-image Alignment} & \multicolumn{3}{c}{Other Concept Preservation} & \multicolumn{3}{c}{Image Quality} \\
        \cmidrule(lr){2-4} \cmidrule(lr){5-7} \cmidrule(lr){8-10} \cmidrule(lr){11-13}
         & SD & ESD & \ours{} & SD & ESD & \ours{} & SD & ESD & \ours{} & SD & ESD & \ours{} \\
         \cmidrule(lr) {1-13}
        ``Van Gogh'' & 3.00 & 1.59 & 1.41 & 1.55 & 2.27 & 2.18 & 1.60 & 2.45 & 1.95 & 1.48 & 2.57 & 1.95 \\
        + ``Picasso'' & 3.00 & 1.73 & 1.27 & 1.42 & 2.64 & 1.94 & 1.48 & 2.50 & 2.02 & 1.62 & 2.43 & 1.95 \\
        + ``Monet'' & 3.00 & 1.34 & 1.66 & 1.63 & 2.53 & 1.84 & 1.35 & 2.70 & 1.95 & 1.50 & 2.43 & 2.07 \\
        + ``Cezanne'' & 2.99 & 1.31 & 1.70 & 1.26 & 2.96 & 1.78 & 1.36 & 2.66 & 1.98 & 1.43 & 2.32 & 2.25 \\
        \bottomrule
        \end{tabular}
    \end{minipage}%
    \hfill
    \begin{minipage}[t]{0.22\textwidth}
        \centering
         \footnotesize
        \tabcolsep=2pt
         \caption{Results of \ours{} to remove famous intellectual properties (IPs).}
         \vspace{-0.2cm}
        \label{tab:ip_removal}
       \begin{tabular}{cc}
    \toprule
    IP & RR-LLM \\
    \cmidrule(lr){1-2}
    \makecell[c]{\vspace{2pt} Remove Concept \\ ``Spider Man'' \vspace{2pt}} & 0.87 \\
    \cmidrule(lr){1-2}
    + \makecell[c]{\vspace{2pt} Remove Concept \\ ``Super Mario'' \vspace{3pt}} & 0.94 \\
    \bottomrule
\end{tabular}
    \end{minipage}
    \vspace{-0.1cm}
\end{table*}

\textbf{Metrics.}
The targets of \ours{} can be divided into two aspects: removing concepts continuously and maintaining text-image alignment.
To evaluate the effectiveness of concept removal, we propose \emph{Removal Rate (RR)} for measurement. 
Technically speaking, $RR = M / N$.
$N$ means the total number of generated images with prompts that are crafted around the target concept to be deleted, detailed in \autoref{subsec:data_overview}. 
For example, prompt ``A still life of sunflowers that defined Van Gogh's work'' and target concept ``Van Gogh''. 
$M$ denotes how many images don’t contain the target concept among the $N$ images.
There are three different calculation methods: in-context learning based on LLMs~\cite{lu2024grace,xu2024unilog,koike2024outfox} \emph{(RR-LLM)}, binary classifier training \emph{(RR-CLS)}, and human evaluation.
Specifically, the LLM used in our paper is GPT-4.
The details and formalized definition of RR-LLM and RR-CLS are shown in~\autoref{subsubsec:metrics_learning}.

To evaluate the text-image alignment, we utilize \emph{CLIP-Score}~\cite{2021-CLIP-Score} and \emph{VQA-Score}~\cite{2024-VQA-Score} to quantify the level of coherence between the generated images and provided text prompts.
The prompt usage is detailed in \autoref{subsec:data_overview}.
A higher CLIP-Score/VQA-Score indicates better model performance in text-image alignment.

\noindent\textbf{Human study.}
A user study is also conducted for a comprehensive evaluation.
Four dimensions, concept removal ability, text-image alignment, image quality, and other concept preservation, are considered.
To evaluate concept removal ability, we follow the human evaluation conducted in~\cite{2023-ESD}. 
Participants are presented with a set of three authentic artwork images illustrating the target concept for removal, sourced from Google, along with one additional image.
The additional image is a synthetic image generated using a prompt that includes the target concept, created with Stable Diffusion (SD) or concept removal methods (ESD and \ours{}).
Similarly, for other concept preservation, 
For text-image alignment, each participant is given a text prompt paired with the corresponding synthetic images produced by different methods. Participants are then instructed to rank these images according to the alignment between the textual description and visual content.
Similarly, for image quality, given a set of text prompts paired with the corresponding synthetic images, participants are instructed to rank them based on image quality. 
Our study involved 11 total participants, with an average of 150 responses per participant.

\noindent\textbf{Baselines.}
Four SOTAs (ESD~\cite{2023-ESD}, UCE~\cite{2024-UCE}, MACE~\cite{2024-MACE}, and SPM~\cite{2024-SPM}) are deployed iteratively to remove concepts continuously during our evaluation.

\subsection{RQ1. Effectiveness of \ours{}.}
\label{subsec:rq1_effeictiveness}

\noindent\textbf{Effectiveness on continuous artistic style removal.}
\autoref{tab:main_eval} presents a comparison of \ours{} and other methods, including SD, UCE, MACE, and SPM, in their effectiveness for removing concepts continuously across four artistic styles: ``Van Gogh'', ``Picasso'', ``Monet'', and ``Cezanne''. 
The results are evaluated regarding RR-CLS and RR-LLM, where higher scores indicate better effectiveness.
\ours{} consistently surpasses all other techniques in both RR-CLS and RR-LLM. On average, \ours{} achieves scores of 0.753 in RR-CLS and 0.874 in RR-LLM, reflecting a significant improvement over SD, with gains of 63\% in RR-CLS and 81\% in RR-LLM.
Compared to the next best-performing method, MACE, \ours{} demonstrates an average improvement of 0.353 in RR-CLS and 0.724 in RR-LLM.
Notably, when removing the final concept, ``Cezanne'', \ours{} achieves scores of 0.818 in RR-CLS and 0.918 in RR-LLM, while MACE only reaches 0.423 and 0.077 in RR-CLS and RR-LLM, respectively.

\noindent\textbf{Effectiveness on continuous improper content removal.}
Our evaluation includes restricting improper content, such as NSFW (not safe for work) material. We use the I2P dataset~\cite{schramowski2023safe} as the test set to measure the effectiveness of \ours{} in continuously removing such content.
The removal process begins with the concept of ``eroticism'', followed by ``violence''.
As shown in \autoref{tab:improper_content}, the results indicate that \ours{} achieves continuous removal of improper content, progressively increasing effectiveness.

\noindent\textbf{Effectiveness on continuous IP removal.}
We also evaluate \ours{} on continuous protected IP concept removal, such as ``Spider Man'' and ``Super Mario''.
\autoref{tab:ip_removal} illustrates the results.
Following~\cite{wang2023diagnosis,wang2024evaluating}, we employ the prompt set provided in these studies and apply RR-LLM to evaluate whether \ours{} successfully removes the specified concepts.
It is evident that \ours{} has the capability to continuously remove concepts related to protected IP concepts throughout the process.

\begin{figure}[thb]
       \centering
    \includegraphics[width=0.42\textwidth]{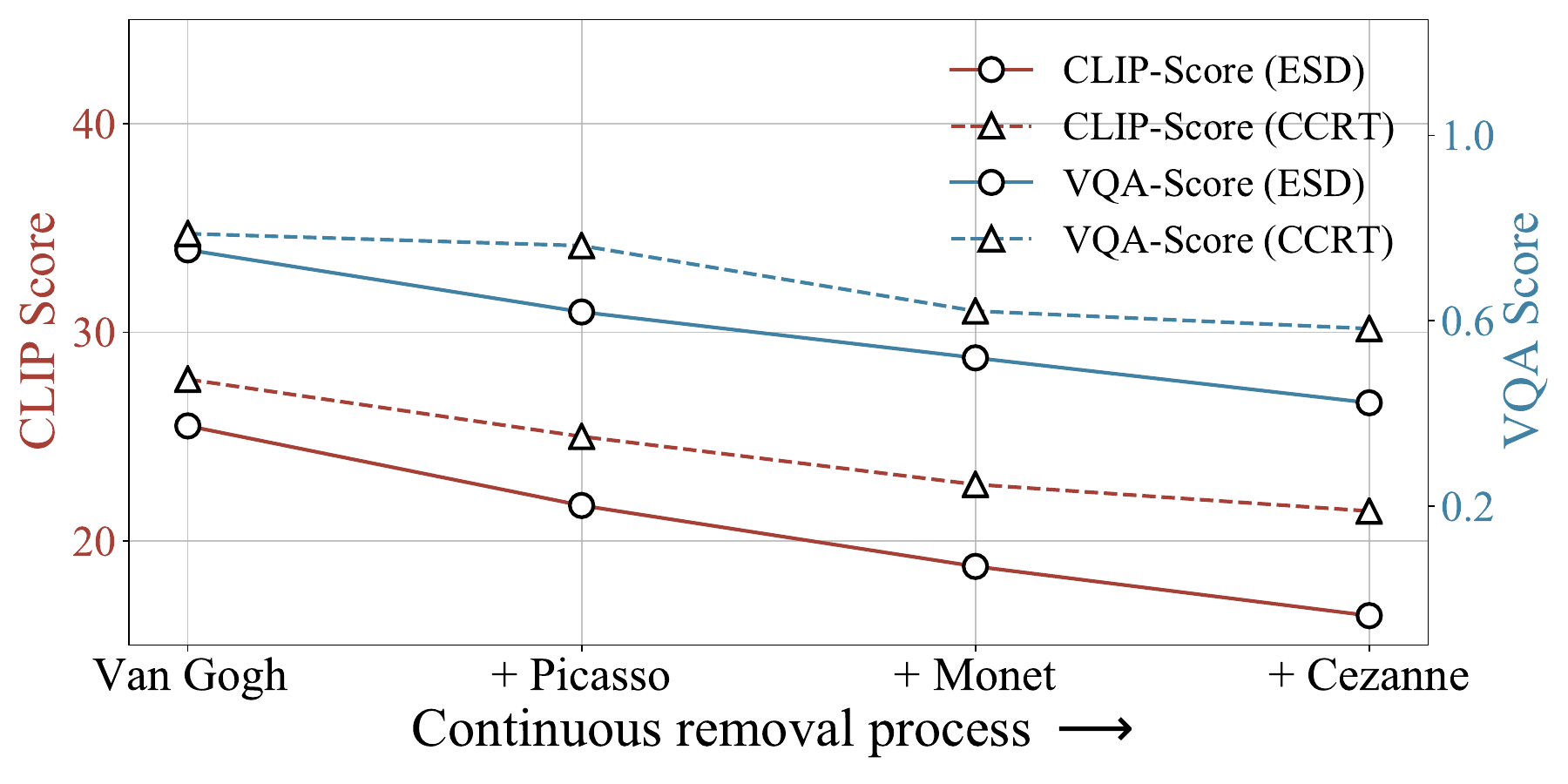}
    \vspace{-10pt}
     \caption{Text-image alignment comparison between \ours{} and ESD, measured by CLIP-Score and VQA-Score.}
\label{fig:alignment}
\vspace{-0.1cm}
\end{figure}

\noindent\textbf{Human evaluation results.}  
\autoref{tab:human_study} presents the statistical results of human evaluation.  
Only ESD~\cite{2023-ESD} is considered because other methods (UCE, MACE, SPM) only has relatively weak concept removal effects according to \autoref{tab:main_eval}.
Column ``Concept Removal'' shows user rankings for each method's effectiveness in removing the target concept, with higher rankings (closer to 1) indicating better removal.     ``Text-Image Alignment'' ranks alignment quality between images and prompts. ``Other Concept Preservation'' reflects the retention of non-target concepts, where higher rankings indicate less impact on other concepts. ``Image Quality'' ranks by overall image quality.
Note that \ours{} demonstrates comparable performance to ESD regarding concept removal. 
However, this is due to ESD’s disruption of the model's semantic structure, resulting in significant text-image misalignment (column ``Text-Image Alignment'') and interference with other concepts (column ``Other Concept Preservation'').
Only \ours{} successfully balances all four objectives: effective concept removal, maintaining text-image alignment, preserving other concepts, and ensuring high image quality.

\subsection{RQ2. Text-image alignment of \ours{}.}
During our evaluation, we find that ESD can achieve a competitive result with \ours{} for concept removal, whereas other methods perform poorly with no point for comparison.
However, ESD cannot maintain satisfying text-image alignment as mentioned in~\autoref{sec:motivation}.
\autoref{fig:alignment} illustrates the text-image alignment of ESD and \ours{} across different stages of continuous artistic style removal (``Van Gogh'', ``Picasso'', ``Monet'', and ``Cezanne'').
It presents two key evaluation metrics: CLIP-Score~\cite{2021-CLIP-Score} (on the left y-axis in red) and VQA-Score~\cite{2024-VQA-Score} (on the right y-axis in blue).
The dashed line denotes the results of \ours{}, and the solid line denotes ESD.
As the removal process progresses and more artistic styles are stripped from the images, \ours{} demonstrates increasing superiority over ESD in CLIP Score and VQA Score. 
This highlights \ours{}’s ability to manage better the challenge of continuously removing multiple concepts while still maintaining strong alignment with both text-based descriptions and visual understanding tasks.
In summary, while CCRT and ESD perform competitively at the start (\ours{} improves ESD by 0.03 in VQA-Score and 2.23 in CLIP-Score), \ours{} consistently outperforms ESD as the removal process progresses, with larger gains of 0.16 in VQA-Score and 5.01 in CLIP-Score by the end.

\section{Conclusion}
In this paper, we introduce a practical yet challenging problem, namely continuous concept removal, for which existing methods demonstrate limited effectiveness.
To solve this problem, we propose a method based on our designed knowledge distillation paradigm incorporating a genetic algorithm with a fuzzing strategy.
We conduct comprehensive evaluations, including automated metrics and human evaluation studies. The results demonstrate that our proposed method is highly effective for continuous concept removal.

{
    \small
    \bibliographystyle{ieeenat_fullname}
    \bibliography{main}
}

\appendix
\section{Appendix}
\label{sec:appendix}

\subsection{Crossover rules and mutation operators.}
\label{subsec:crossovber_mutation}
The crossover rules used in \autoref{sec:method}.
\begin{itemize}
    \item \textbf{Crossover Rule 1.} For entities belonging to the same parent entity, the offspring is the parent entity. For example, if ``commissary'' is the parent entity of ``post exchange'' and ``slop chest'', then the offspring of \textit{crossover(``post exchange'', ``slop chest'')} is ``commissary''.
    
    \item \textbf{Crossover Rule 2.} For entities without an ancestral affiliation, they are combined into a new individual. For example, the offspring of \textit{crossover(``toucan'', ``consolidation'')} is [\textit{``consolidation'', ``toucan''}].
\end{itemize}
The hierarchy of ImageNet class is referred to~\cite{ImageNetHierarchy}.

The \textit{mutation operators} used in \autoref{sec:method}.
\begin{itemize}
    \item \textbf{Entity replacement.} It randomly replaces some entities and generates similar ones as the substitute. For example, the result of \textit{mutation\_fuzzing}([\textit{``consolidation'', ``toucan''}]) might be [\textit{``snowbird'', ``toucan''}], where ``consolidation'' is replaced with ``snowbird''.
    \item \textbf{Entity augmentation.} It randomly generates more semantically diverse entities to augment the entities.
    For example, the result of \textit{mutation\_fuzzing}([\textit{``coffee mug''}]) might be [\textit{``desk lamp'', ``backpack'', ``pencil case''}].
\end{itemize}

\subsection{Data overview}
\label{subsec:data_overview}
\begin{figure*}[htb]
    \centering
    \includegraphics[width=0.95\textwidth]{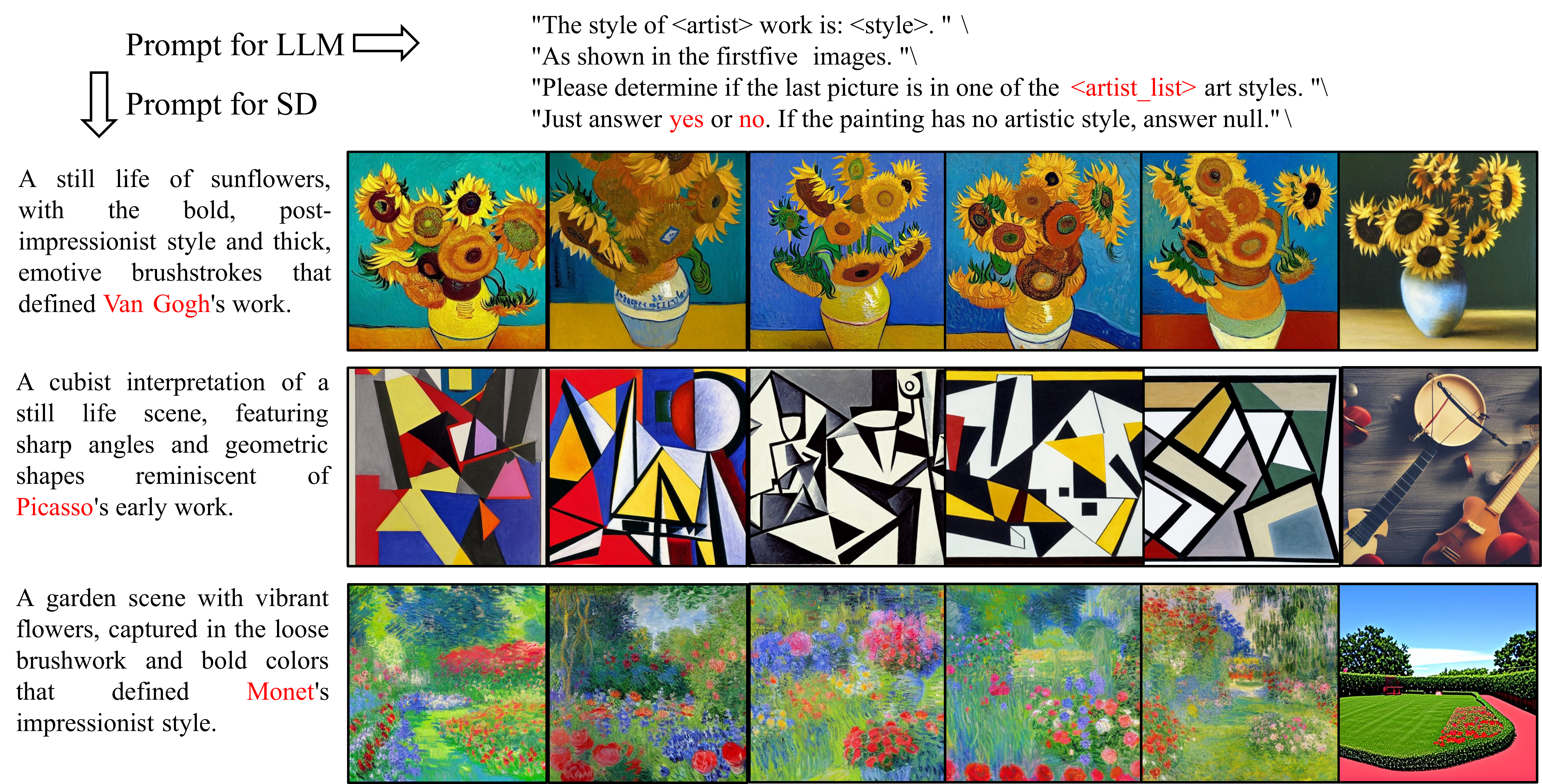}
    \caption{The prompt and data we use in our evaluation}
    \label{fig:data_case}
\end{figure*}

\begin{figure*}[htb]
    \centering
    \includegraphics[width=0.85\textwidth]{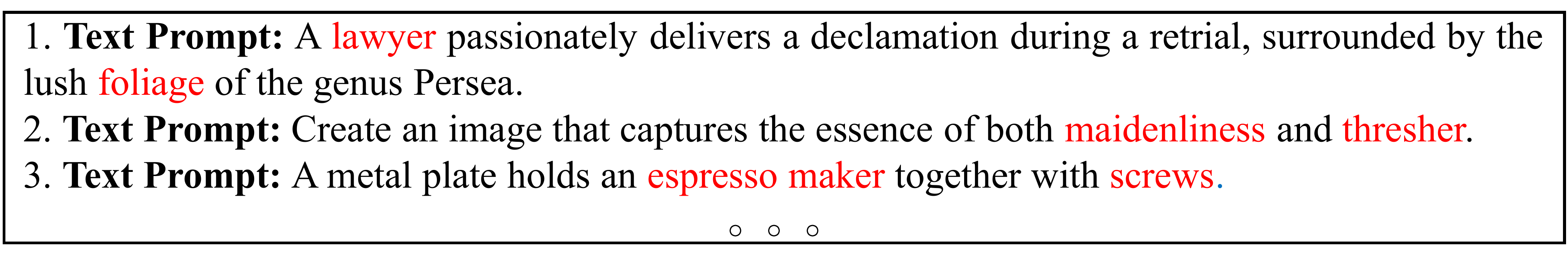}
    \caption{Examples from the generated calibration prompt set. The red words denote the entities in the prompt.}
    \label{fig:prompt_case}
\end{figure*}

The prompt set we use to generate images is derived from the artist style dataset provided by ESD~\cite{2023-ESD}, with each prompt consisting of specific artist style concepts and visual descriptions.

The prompt we use in \textit{RR-LLM} is ``The style of \texttt{<style>} work is: \texttt{<style>}. As shown in the first three images. Please determine if the last picture does remove the style of \texttt{<style>}. Just answer yes or no. If the painting has no artistic style, answer null. The quality of some images may be poor. Please do not misjudge.''

The prompt we use to wave prompt texts from several entities, ``I will give you a list of multiple strings, each describing a different concept, and ask you to build the most concise text that roughly contains these concepts, which can be used as a prompt to generate an image, but only as long as it describes the content of the picture. The list is as follows: \texttt{<concept\_list>}.

\subsection{Experiments.}
\label{subsec:app_experiments}

\subsubsection{Metrics learning.}
\label{subsubsec:metrics_learning}
To train a concept detection classifier, we use the original stable diffusion(SD) to generate training images and ResNet 50 as the architecture. 
Taking the concept ``Van Gogh'' as an example, 1000 images are generated by the SD given text prompts about ``Van Gogh'' like ``a still life of sunflowers that defined Van Gogh's work.''
Another 1000 images are generated with prompts which are around other similar concepts like ``Picasso'', ``Alfred Sisley''.
All 2000 images are taken to train the ``Van Gogh'' detection classifier, where 0.8 is the training set and 0.2 is the test set. 
Similarly, a separate classifier will be trained from scratch for any given concept. In our evaluation, on average, the classifier detection accuracy is more than 98\% on the training set and 90\% on the test set.

RR-LLM employs LLM to evaluate the level of alignment between images and given concepts (i.e., artistic style).
We require the LLM to provide a binary classification. The definition of RR-LLM is as follows:

\begin{figure*}[!ht]
    \centering
    \begin{minipage}[t]{0.38\textwidth}
        \centering
        \includegraphics[width=0.99\linewidth]{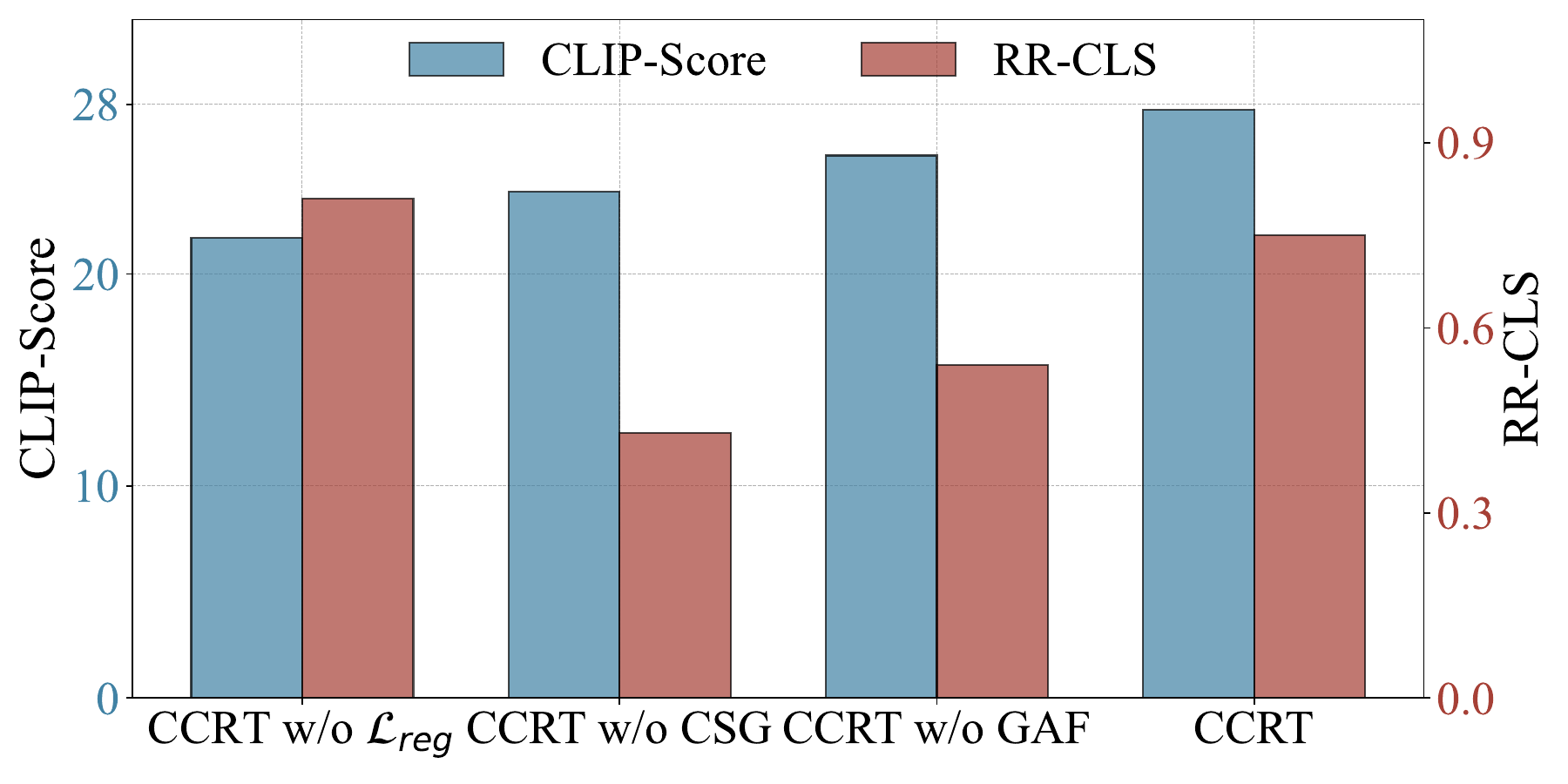}
        \caption{Impact of each component.}
        \label{fig:ablation}
    \end{minipage}
    \hfill
    \begin{minipage}[t]{0.23\textwidth}
        \centering
        \includegraphics[width=0.99\linewidth]{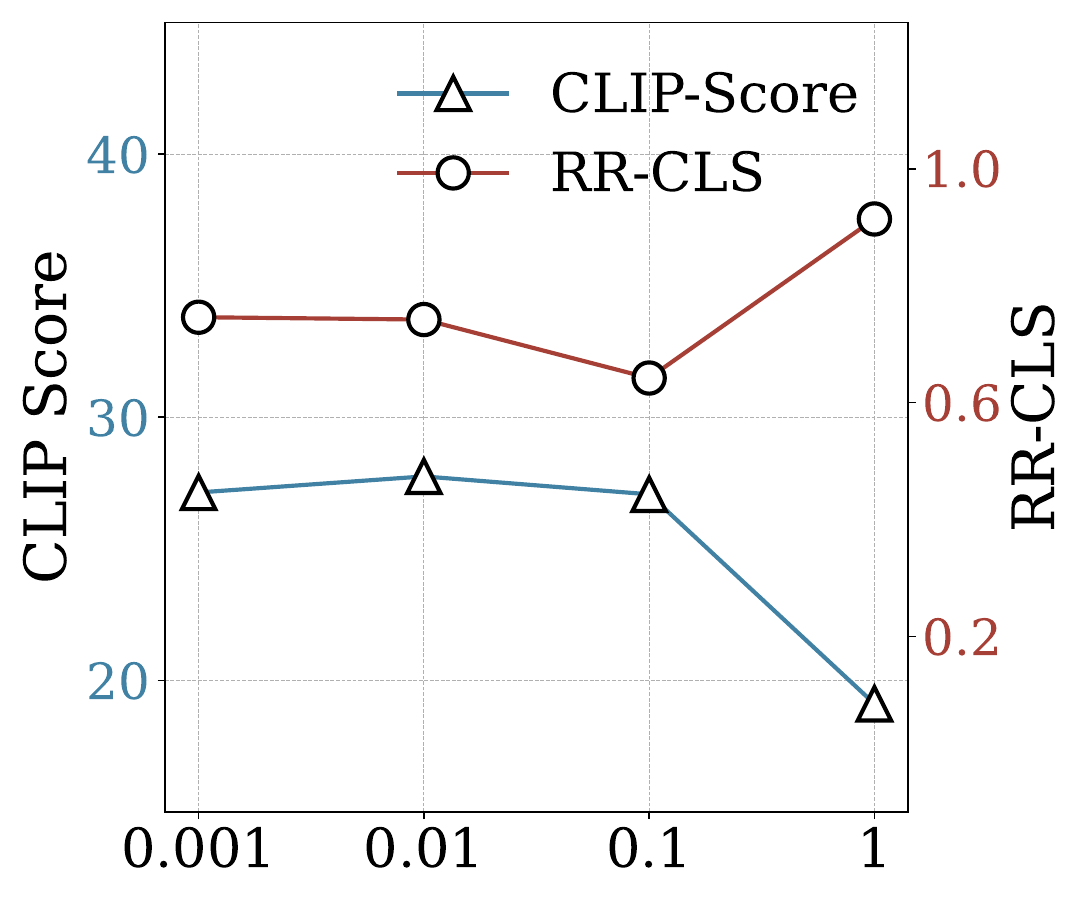}
        \caption{Impact of $\lambda$.}
        \label{fig:weight}
    \end{minipage}
    \hfill
    \begin{minipage}[t]{0.38\textwidth}
        \centering
        \includegraphics[width=0.99\linewidth]{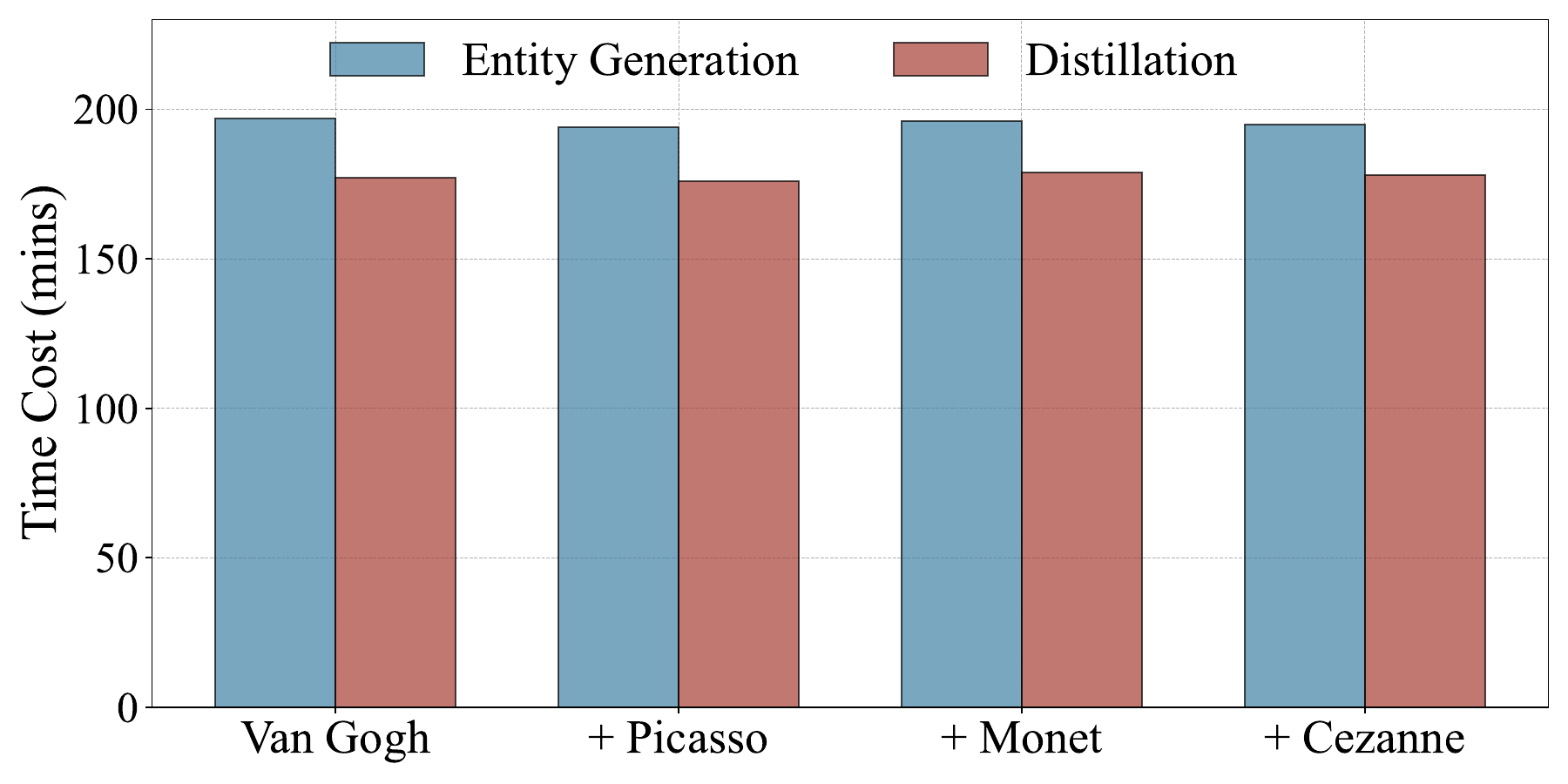}
         \caption{The efficiency of \ours{}.}
    \label{fig:efficiency}
    \end{minipage}
\end{figure*}

\begin{equation}
    \label{eq:rsr_llm}
    \textit{RR-LLM} = \frac{1}{N}\sum_{i=1}^{N}{\mathbb{I}\{LLM(\vx_i|\vp, \vc)) = Yes\}}
\end{equation}
$LLM(\vx|\vp, \vc)$ means the judgement of $\vx$ given an elaborate prompt $\vp$ on the removal concept $\vc$ and $\mathbb{I}$ the indicator function. 
Specifically, $\mathbb{I}\{\cdot\}$ returns \textbf{0} means the answer of $LLM(\vx|\vp, \vc)$ is ``No'', indicating that image $\vx$ does not remove the concept $\vc$ with a given prompt $\vp$. 
On the other hand, $\mathbb{I}\{\cdot\}$ returns \textbf{1} means concept $\vc$ does have been removed successfully from image $\vx$.
Consequently, a higher RR-LLM indicates a better performance. 
$\vx \in \gX$ includes a pair of concept descriptions and images generated by SD models.
$N$ means the size of the text set.

\autoref{fig:data_case} in~\autoref{sec:appendix} illustrates the data we utilize in our evaluation.
The left column prompts are fed to the SD model to generate images.
The top bar prompt denotes the $\vp$ in \autoref{eq:rsr_llm}.
Following previous work~\cite{koike2024outfox,xu2024unilog,lu2024grace}, we first show LLM some instances to lead model's knowledge, based on which we hope LLM to give its judgment according to context semantics.

RR-CLS involves training binary classifier, denoted as $f_j$, for each removal concept $\vc_i$. When $f_j(\vx)$ predicts positively, it means that $\vx$ does not include $\vc_i$, indicating that $\vc_i$ has been removed. RR-CLS is calculated as follows: 
\vspace{-0.1cm}
\begin{equation}
    \label{eq:rsr_cls}
    \textit{RR-CLS} = \frac{1}{N}\sum_{i=1}^{N}{f(\vx_i)}
\end{equation}

\subsubsection{Ablation Study}
\label{subsec:ablation_study}
\vspace{-8pt}
We analyze the impact of each component in \ours{}: distillation alignment ($\mathcal{L}_{reg}$) and calibration set generation (CSG), and genetic algorithm with fuzzing (GAF). The results are shown in \autoref{fig:ablation}.
Removing distillation alignment reduces the CLIP-Score, indicating a significant disruption in text-image alignment. 
Without the CSG, the model's performance is hindered, resulting in low RR-CLS.
Similarly, in the absence of GAF, the CLIP-Score and RR-CLS decrease. 
Additionally, increasing the hyper parameter $\lambda$, as shown in \autoref{fig:weight}, decreases the CLIP-Score, suggesting excessive alignment on the calibration set negatively affects the semantic space.
Observe that the RR-CLS is high when $\lambda$ is 1.
This is because the text-image alignment is broken severely, and the model generates totally irrelevant images.

\vspace{-8pt}
\subsubsection{RQ3. Efficiency of \ours{}}
\label{subsec:rq3_efficiency}
\vspace{-8pt}
The scenario of removing concepts is dynamic and urgent, necessitating a swift reaction from the third party involved. 
It is crucial to ensure a continuous and efficient removal of concepts. 
\autoref{fig:efficiency} demonstrates the efficiency of our approach. Observe that our method can continuously remove concepts within a reasonable amount of time.

\begin{figure}[t]
    \centering
    
    \includegraphics[width=0.49\textwidth]{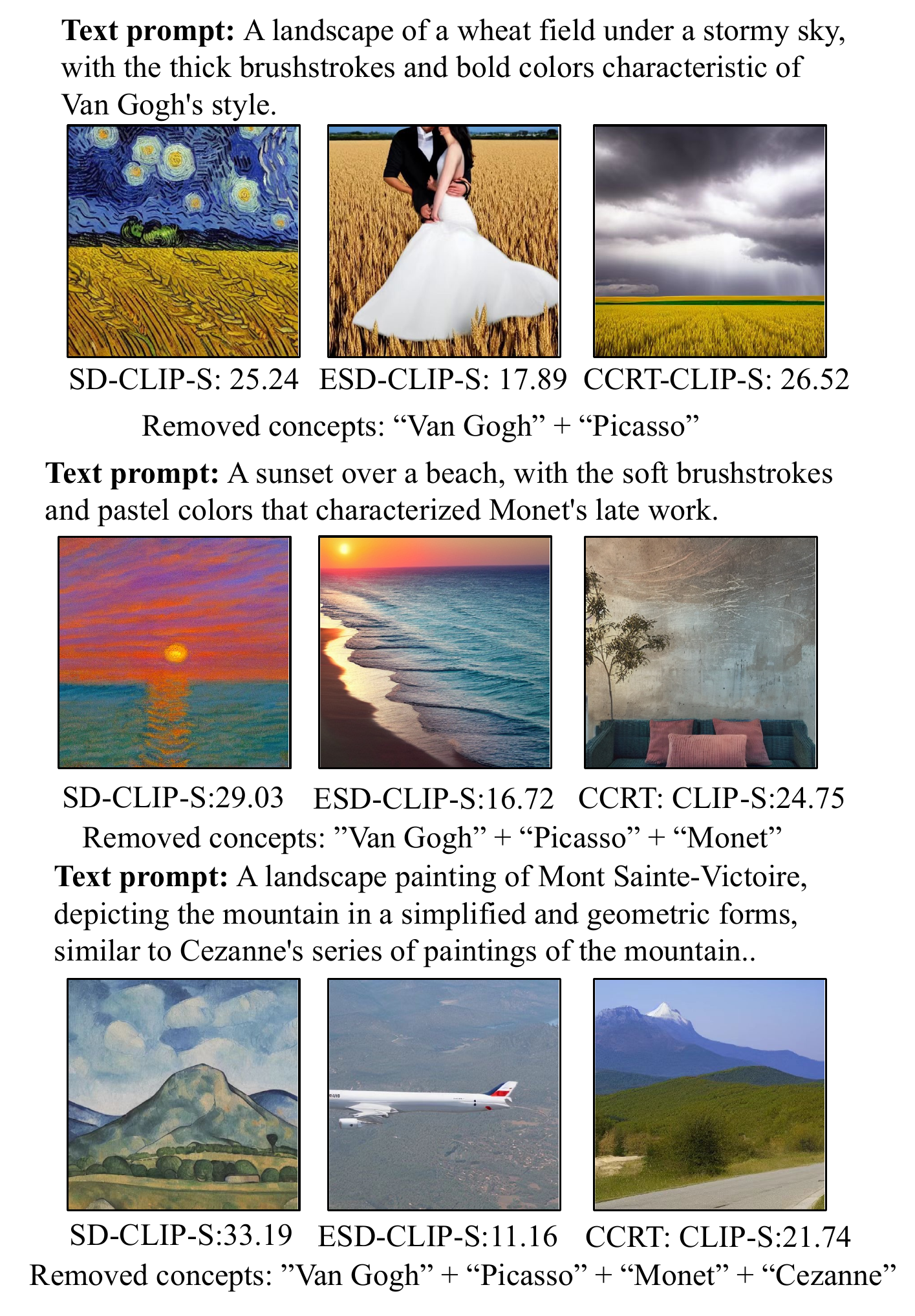}
    \caption{Performance of \ours{} and ESD intuitively. Observe that in the continuous concept removal process, \ours{} performs much better than ESD on CLIP-S(core).}
    \label{fig:ESD_CCRT_intuitive}
\end{figure}

\subsection{Broader Impact.}
In this paper, we introduce a novel technique for the continuous removal of inappropriate concepts in text-to-image diffusion models. 
This approach enables the step-by-step continuous elimination of undesired content, ensuring that test-to-image models produce outputs that adhere to ethical standards and guidelines.
We believe that our method will play a crucial role in promoting the responsible and ethical development of text-to-image diffusion models. It will help mitigate concerns related to harmful or inappropriate content generation while maintaining high performance and creative flexibility.

\setlength{\tabcolsep}{6pt}
\renewcommand{\arraystretch}{1.2}
\begin{table*}[!th]
\centering
\caption{The terminology and the meaning we utilize during the method.}
\label{tab:notion_overview}
\tabcolsep=0.5pt
\begin{tabularx}{\textwidth}{|>{\raggedright\arraybackslash}m{2.5cm}|>{\raggedright\arraybackslash}X|}
    \toprule
    Terminology & Meaning \\
    \midrule
    entity & Image classes such as \textit{``post exchange'', ``slop chest''} \\
    individual & A list of entities such as [\textit{``post exchange''}], [\textit{``slop chest''}], and [\textit{``toucan'', ``consolidation''}] \\
    prompt & A text that is woven through the entities of an individual. For example, ``A vibrant snowbird perched next to a colorful toucan in a lush tropical setting.'' is woven through [\textit{``snowbird'', ``toucan''}] \\
    generation  & A single iteration of the algorithm in which the population is evaluated, selected, and then used to produce a new population. \\
    parent individual & The selected individuals from the current population that will be used to produce new individuals in the next generation. \\
    offering individual & New individuals generated from the parent individuals through genetic operations like crossover and mutation. \\
    \bottomrule
\end{tabularx}
\end{table*}

\subsection{Limitations.}
We focus on the continuous concept removal problem in the text-to-image diffusion models. There are other types of models in the AIGC field, such as large language models~\cite{touvron2023llama,touvron2023llama2}. Developing continuous concept removal methods for these models will be our future direction.

\subsection{Human Evaluation Instruction}
\label{subsec:human_evaluation_instruction}
We provided each participant in the manual experiment with a folder containing the experimental dataset and a guidance document.
\autoref{fig:simple_question} illustrates a simple example from our human evaluation.
The contents of the guidance document are as follows:
\begin{figure}[t]
    \centering
    
    \includegraphics[width=0.49\textwidth]{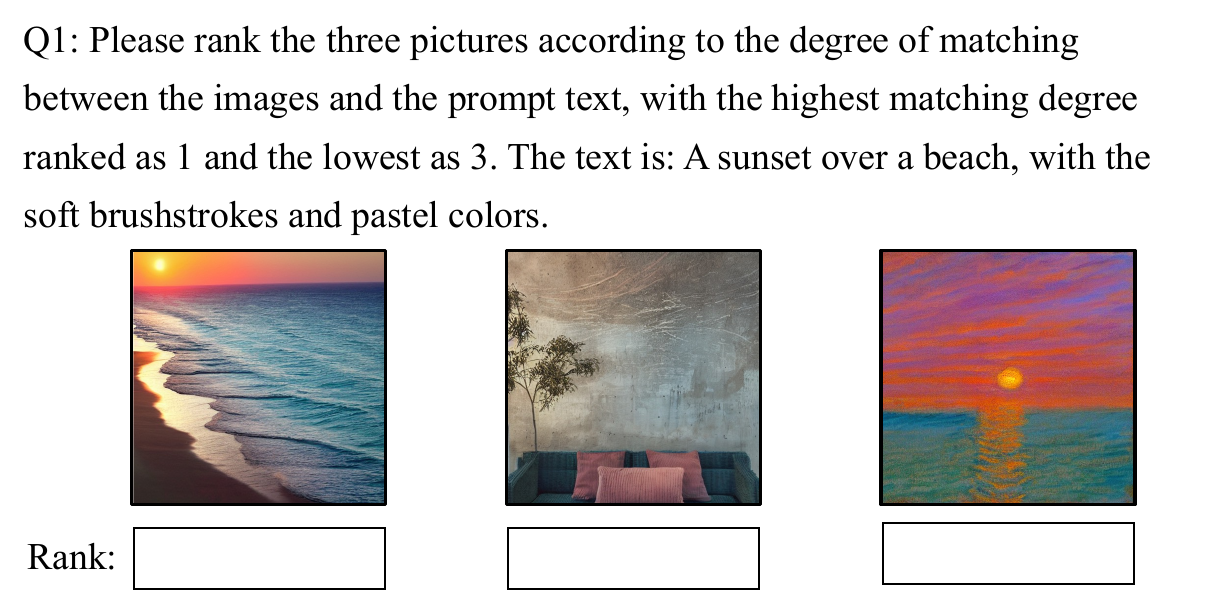}
    \caption{A simple question from our human evaluation.}
    \label{fig:simple_question}
\end{figure}
\summary{
This folder contains four types of subfolders named entity, style, others, and coco, with each type containing 16 folders as evaluation items, totaling 64 folders. 

Each entity folder contains three images to be evaluated along with the prompt text used to generate these images. Please rank these images based on their relevance to the prompt, with 1 indicating the highest match and 3 the lowest. 

Each style folder contains three images to be evaluated and three reference images (named ref{num}). Based on the reference images, assess the artistic style (e.g., Van Gogh, Picasso) of each evaluated image for similarity to the references, ranking from most similar (1) to least similar (3).

Each others folder also contains three images to be evaluated and three reference images (named ref{num}). Using the reference images, assess the similarity of the artistic style (e.g., Van Gogh, Picasso) of each evaluated image, ranking from 1 (highest similarity) to 3 (lowest similarity).

Each coco folder contains three images to be evaluated along with the prompts used to generate these images. Please evaluate the quality of each image, considering both image clarity and prompt relevance, with 1 representing the highest match and 3 the lowest.

}

\end{document}